\documentclass[conference]{IEEEtran}
\IEEEoverridecommandlockouts
\usepackage{cite}
\usepackage{amsmath,amssymb,amsfonts}
\usepackage{algorithmic}
\usepackage{graphicx}
\usepackage{textcomp}
\usepackage{xcolor}
\def\BibTeX{{\rm B\kern-.05em{\sc i\kern-.025em b}\kern-.08em
    T\kern-.1667em\lower.7ex\hbox{E}\kern-.125emX}}

\makeatletter
\newcommand{\linebreakand}{%
  \end{@IEEEauthorhalign}
  \hfill\mbox{}\par
  \mbox{}\hfill\begin{@IEEEauthorhalign}
}
\makeatother

\begin{document}

\title{CodeForge-MA: Execution-Verified Multi-Agent Learning with Language-Conditioned LoRA for Multilingual Code Generation\\}

\author{
\IEEEauthorblockN{Zhizhou Gu*}
\IEEEauthorblockA{\textit{Northeastern University} \\
Bellevue, USA \\
gu.zhiz@northeastern.edu}
\and
\IEEEauthorblockN{Xianting Wu}
\IEEEauthorblockA{\textit{Columbia University} \\
New York, USA \\
xianting.wu@outlook.com}
\and
\IEEEauthorblockN{Siyu Gu}
\IEEEauthorblockA{\textit{University of Pennsylvania} \\
Philadelphia, USA \\
gusiyu88@engineering.upenn.edu}
\and
\linebreakand
\IEEEauthorblockN{Tian Zhang}
\IEEEauthorblockA{\textit{University of Southern California} \\
San Jose, USA \\
0tianzhang0@gmail.com}
\and
\IEEEauthorblockN{Kejian Tong}
\IEEEauthorblockA{\textit{Independent Researcher} \\
Mukilteo, USA \\
tongcs2021@gmail.com}
}

\maketitle

\begin{abstract}
Large language models for code generation often fail on execution, multilingual coverage, and contamination control, especially under frozen backbone constraints. We present CodeForge-MA, a unified framework that improves code synthesis through a multi-agent data forge, execution verified reinforced instruction tuning, and a language conditioned mixture of LoRA adapters. Four specialized agents, Composer, Reviewer, Executor, and Curator, iteratively refine instruction code pairs, validate them with tests, and filter duplicates and benchmark leakage. During training, we combine masked supervised fine tuning with a test driven reinforcement objective to align generations with executable correctness. For the larger model, we use sparse expert routing over low rank adapters to improve cross language transfer while keeping the base model unchanged at inference. Experiments show that joint data, objective, and adapter design yields robust gains across programming languages.
\end{abstract}


\section{Introduction}
Generating correct code from natural language remains a critical challenge because success requires more than fluent text production, it demands syntactic validity, semantic correctness, and robust behavior across diverse programming languages. In practice, many systems produce answers that appear plausible but fail at compilation or unit testing, which makes execution grounded evaluation indispensable for real code tasks \cite{wei2022chain}.Analogous reliability concerns arise in high-stakes LLM deployments such as financial anti-fraud systems, where hallucinations and unverifiable numerical or logical claims motivate layered verification and guardrail mechanisms \cite{huo2026suppressing}.

Recent advances in prompting and large language model reasoning have improved code synthesis, yet they still struggle with long tail languages, prompt contamination, and the mismatch between reference style and executable correctness. Instruction tuning alone often optimizes surface form, while test time search and reasoning strategies add cost without directly fixing data quality or adapter efficiency \cite{wang2022self}. At the same time, scaling model size is not always feasible when the backbone must remain frozen and inference overhead must stay low \cite{rafailov2023direct}.These efficiency concerns are especially acute for edge-class Arm CPU deployment, where memory capacity, bandwidth pressure, low-precision throughput, and kernel-level optimization jointly constrain practical LLM inference \cite{zhou2026roofline}.This emphasis on architecture-level efficiency under realistic deployment constraints is consistent with recent evidence that different LLM reasoning architectures can exhibit markedly different accuracy, auditability, and cost trade-offs in resource-constrained settings \cite{liu2026architecture}.

To address these constraints, we introduce CodeForge MA, a framework that jointly improves training data, optimization objective, and parameter efficient adaptation. Our approach uses a multi agent data forge to create and verify code samples, an execution verified reinforced instruction tuning objective to reward passing tests, and a language conditioned mixture of low rank adapters to specialize efficiently across programming languages. The result is a frozen backbone pipeline that concentrates learning where it matters most, in the data, the reward signal, and the adapters.

\section{Related Work}
Work on core code generation has established that large language models can synthesize programs from natural language, but benchmark performance remains sensitive to dataset quality, language imbalance, and execution reliability \cite{chen2021evaluating}. Recent studies also show that model scaling alone does not resolve compilation errors or hidden test failures, especially in multilingual settings where training corpora are unevenly distributed \cite{kocetkov2022stack}.

Auxiliary approaches have therefore focused on strengthening supervision through better prompting, self refinement, and execution feedback. Instruction tuning and reasoning centered decoding can improve outputs, yet they still depend heavily on the quality of the underlying synthetic or curated data, and they may not reliably eliminate contaminated examples or silent failures \cite{wang2024lora}. Parameter efficient adaptation methods further reduce training cost, but standard adapters or low rank modules can underutilize capacity when task diversity is high \cite{dettmers2023qlora}.

Our framework is closest to recent efforts that combine data curation, execution grounded learning, and modular adaptation for code oriented LLMs. In contrast to single stage tuning, we integrate a multi agent forge with verification, a reinforced instruction objective tied to unit tests, and language conditioned expert routing over LoRA modules to reduce cross language interference. This design is motivated by broader findings that structured supervision and sparse adaptation can improve controllability and efficiency in large scale language models \cite{fedus2022switch}\cite{du2022glam}.Related ideas about robustness through structured representations and causal disentanglement have also been explored in graph-based fraud detection, where suppressing spurious environmental correlations improves generalization under distribution shift \cite{luo2025robust}.Similar efficiency-driven design principles have also appeared in cloud-native performance prediction, where multimodal generative modeling, long-range dependency mechanisms, and calibrated uncertainty estimation are combined to support online deployment under dynamic workloads \cite{gu2026efficient}.

\section{Methodology}
\begin{figure*}[t]
\centering
\includegraphics[width=\textwidth]{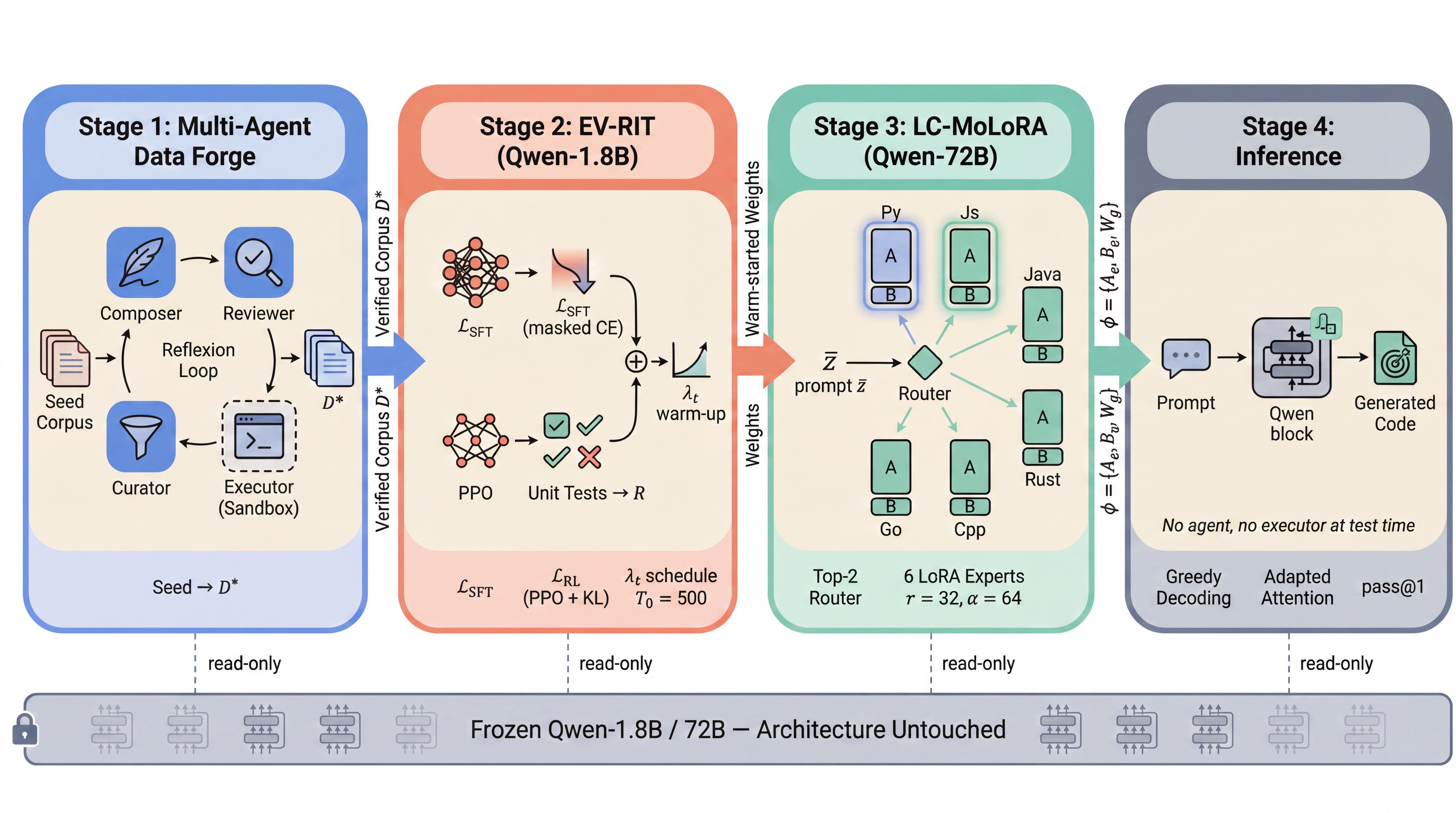}
\caption{Overview of the CodeForge-MA framework. Four sequential stages
  (Multi-Agent Data Forge, EV-RIT Training, LC-MoLoRA Adaptation, Inference)
  operate above a frozen Qwen-1.8B/72B backbone, faithfully respecting the
  competition's no-modification rule.}
\label{fig:163_1}
\end{figure*}

Under a frozen-backbone, limited-compute setting, architectural novelty is unavailable, so gains must come from the data pipeline, the supervised objective, and the parameter-efficient adaptation layer. Built on the Qwen-1.8B and Qwen-72B backbones—decoder-only Transformers with RoPE, RMSNorm, SwiGLU, and tied embeddings—we propose CodeForge-MA, a unified framework pairing a multi-agent data forge with a mixture-of-adapter tuning layer. On the data side, four specialized LLM agents—Composer, Reviewer, sandboxed Executor, and Curator—cooperate in a Reflexion-style loop that drafts instruction-code pairs, critiques them via self-consistency voting, compiles and tests them against auto-generated unit tests, and finally deduplicates and decontaminates survivors against HumanEval, HumanEvalFix, and MBPP using MinHash LSH. This addresses three practical failures we observed: silent compile-time errors in single-shot synthetic data, the long-tail scarcity of Rust and Go samples, and prompt-level leakage in Evol-Instruct-style corpora. On the optimization side, we introduce an execution-verified reinforced instruction-tuning objective that combines token-level cross-entropy on the assistant turn with a PPO-style surrogate whose reward is the unit-test pass ratio, gated by a warm-up schedule so fluency is acquired before execution signals are injected. For the 72B regime, where full fine-tuning is infeasible, we attach a language-conditioned mixture-of-LoRA module to the attention projections, in which six low-rank experts are blended by a top-$k$ router conditioned on the mean-pooled prompt representation, regularized by a Switch-style load-balancing term and a soft language-identity prior to prevent expert collapse and encourage cross-lingual transfer. The Qwen architecture remains strictly untouched at inference, adding only negligible routing overhead, while all trainable capacity is concentrated in the agentic curation process and the low-rank adapters. Fig.~\ref{fig:163_1} summarizes how the multi-agent data forge, the execution-verified reinforced instruction tuning stage, and the language-conditioned mixture-of-LoRA adapter stack on top of the strictly frozen Qwen backbone.

\section{Algorithm and Model}
\label{sec:algorithm}

In this section we describe CodeForge-MA, the end-to-end framework we designed for the Qwen AI Challenge. Because the competition forbids any modification to the backbone network itself, everything that can actually move the needle, i.e.\ the training data, the supervised objective, and the parameter-efficient adaptation, has to be reconsidered jointly rather than in isolation. We therefore organize the method around three tightly coupled components, preceded by a short description of the base model we build upon.

\subsection{Base Model: Qwen-1.8B and Qwen-72B}
\label{subsec:base}

We adopt Qwen-1.8B for the preliminary round and Qwen-72B for the final round, both released by Alibaba Cloud in late 2023 under an open weight license. Architecturally they are decoder-only autoregressive Transformers sharing the same recipe, differing only in depth, width and number of attention heads. Given an input token sequence $\mathbf{x}=(x_1,\dots,x_T)$ drawn from a byte-level BPE vocabulary of roughly $152$K symbols, the model factorizes the joint distribution in the standard causal way,
\begin{equation}
p_\theta(\mathbf{x}) \;=\; \prod_{t=1}^{T} p_\theta\!\left(x_t\,\middle|\,x_{<t}\right),
\label{eq:ar}
\end{equation}
and stacks $L$ pre-norm Transformer blocks equipped with Rotary Position Embedding (RoPE), Root-Mean-Square Normalization and a SwiGLU feed-forward network:
\begin{align}
\tilde{\mathbf{h}}^{(\ell)} &= \mathbf{h}^{(\ell-1)} + \mathrm{MHA}\!\big(\mathrm{RMSNorm}(\mathbf{h}^{(\ell-1)})\big), \label{eq:block1}\\
\mathbf{h}^{(\ell)} &= \tilde{\mathbf{h}}^{(\ell)} + \mathrm{FFN}_{\mathrm{SwiGLU}}\!\big(\mathrm{RMSNorm}(\tilde{\mathbf{h}}^{(\ell)})\big). \label{eq:block2}
\end{align}
The attention in each head reads
\begin{equation}
\mathrm{Attn}(\mathbf{Q},\mathbf{K},\mathbf{V}) = \mathrm{softmax}\!\left(\tfrac{\mathbf{Q}\mathbf{K}^{\!\top}}{\sqrt{d_h}}+\mathbf{M}\right)\mathbf{V},
\label{eq:attn}
\end{equation}
with causal mask $\mathbf{M}$, and the SwiGLU block is
\begin{equation}
\mathrm{FFN}(\mathbf{x}) = \big(\mathrm{SiLU}(\mathbf{x}\mathbf{W}_g)\odot\mathbf{x}\mathbf{W}_u\big)\mathbf{W}_d,
\label{eq:swiglu}
\end{equation}
where $\mathrm{SiLU}(u)=u\,\sigma(u)$. An important practical detail is that the base checkpoint uses $\langle\texttt{endoftext}\rangle$ as its EOS, whereas the fine-tuned ChatML variant switches to $\langle\texttt{im\_end}\rangle$; mixing the two is a subtle but common source of degenerate generations, and throughout our pipeline we explicitly re-tokenize every sample so that the EOS token is consistent with the target format. Qwen also ties input and output embeddings, which we deliberately do not break, to keep the gradient footprint minimal during parameter-efficient tuning.

\subsection{Multi-Agent Data Forge}
\label{subsec:madf}

\begin{figure}[htbp]
\centering
\includegraphics[width=0.5\textwidth]{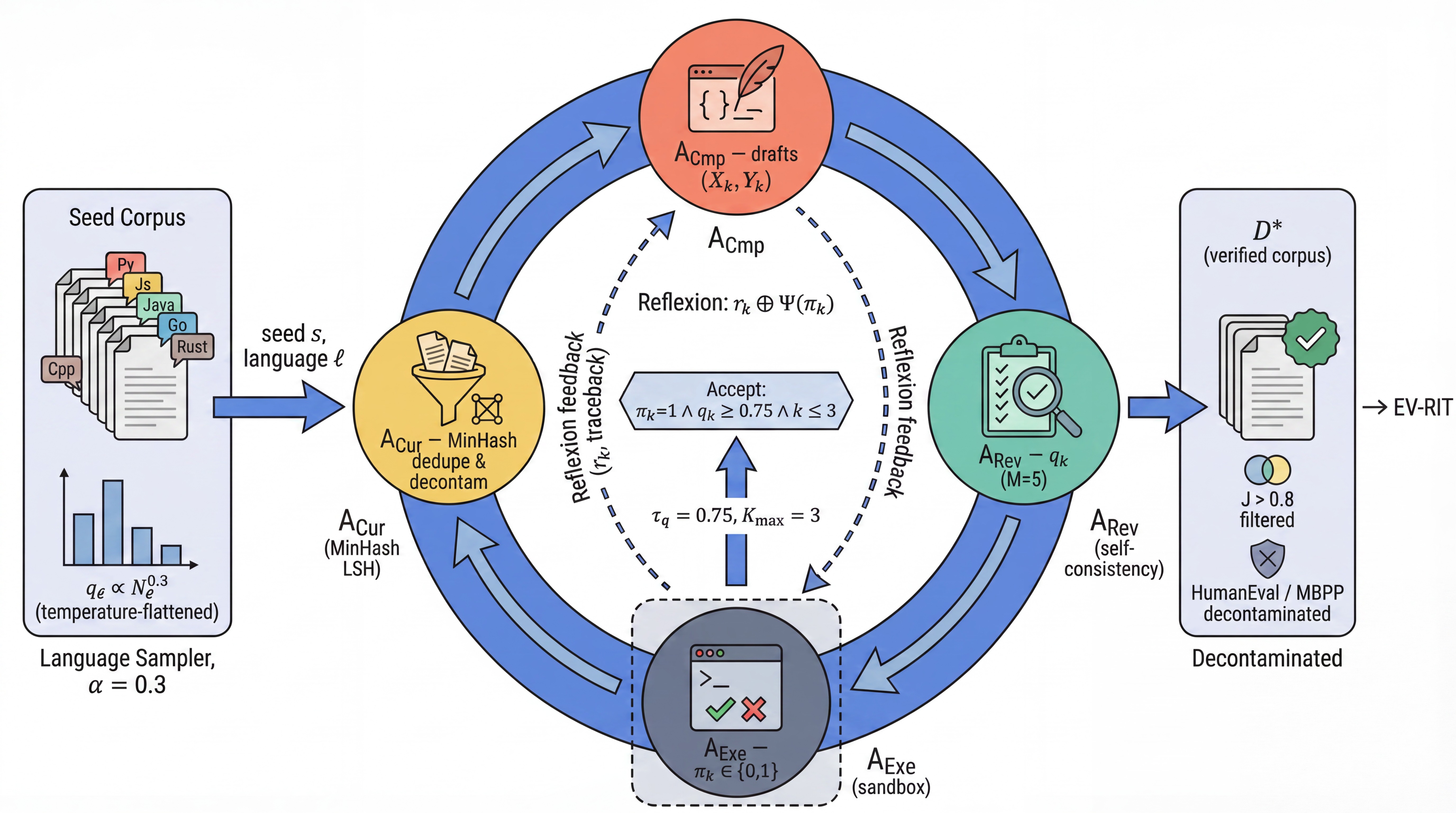}
\caption{The Multi-Agent Data Forge. Four specialized agents
  $\{A_{\mathrm{Cmp}},A_{\mathrm{Rev}},A_{\mathrm{Exe}},A_{\mathrm{Cur}}\}$
  cooperate in a closed Reflexion-style loop; a sample is accepted only when
  $\pi_k=1$, $q_k\ge\tau_q$, and $k\le K_{\max}$.}
\label{fig:163_2}
\end{figure}

The first component of CodeForge-MA is a hierarchical multi-agent data forge. The rationale is that the scarcity of high-quality, execution-verified multilingual code instructions, not model capacity, is the real bottleneck in this competition. Instead of scraping and lightly filtering as the common baselines do, we organize four specialized LLM agents into a closed control loop. The agent set is $\mathcal{A}=\{A_{\mathrm{Cmp}},A_{\mathrm{Rev}},A_{\mathrm{Exe}},A_{\mathrm{Cur}}\}$, where $A_{\mathrm{Cmp}}$ is a Composer that drafts a candidate instruction-code pair from a seed, $A_{\mathrm{Rev}}$ is a Reviewer that scores style, correctness and pedagogical value, $A_{\mathrm{Exe}}$ is an Executor that compiles and runs the code inside a sandbox, and $A_{\mathrm{Cur}}$ is a Curator that performs global deduplication and decontamination. Formally, the state of a sample at iteration $k$ is $z_k=(X_k,Y_k,r_k,q_k,\pi_k)$, where $X_k$ is the prompt, $Y_k$ the code, $r_k$ the natural-language review, $q_k$ a scalar quality score and $\pi_k\in\{0,1\}$ the execution indicator. The forge transitions according to
\begin{equation}
z_{k+1} \;=\; \mathcal{F}(z_k;\mathcal{A}) \;=\; A_{\mathrm{Exe}}\circ A_{\mathrm{Rev}}\circ A_{\mathrm{Cmp}}(z_k),
\label{eq:transition}
\end{equation}
and a sample is accepted only when
\begin{equation}
\mathrm{Accept}(z_k) \;=\; [\pi_k=1]\wedge[q_k\ge\tau_q]\wedge[k\le K_{\max}],
\label{eq:accept}
\end{equation}
with $\tau_q=0.75$ and $K_{\max}=3$. We found empirically that a naive single-pass synthesis protocol produces code which frequently looks plausible but fails to compile, particularly in Rust and Go where trait bounds and package declarations are easy to miss; constraining every survivor through $\pi_k=1$ eliminates this failure mode at the source. The closed-loop structure of the forge, including the Reflexion-style feedback path from the Executor back to the Composer, is depicted in Fig.~\ref{fig:163_2}.

To stabilize the Reviewer, which is itself an LLM and therefore a noisy judge, we apply self-consistency. The Reviewer is sampled $M$ times with temperature $\tau_R$, and the scalar score is
\begin{equation}
q_k \;=\; \frac{1}{M}\sum_{m=1}^{M} A_{\mathrm{Rev}}^{(m)}(X_k,Y_k),\quad M=5,
\label{eq:selfconsist}
\end{equation}
which, in our ablations, correlates more reliably with downstream pass rate than a deterministic single critique. Whenever $\mathrm{Accept}(z_k)$ returns false, we do not discard the sample immediately. Instead, the Composer is re-invoked with a prompt augmented by the Reviewer's report and the Executor's traceback,
\begin{equation}
\mathrm{Prompt}_{k+1} \;=\; \mathrm{Prompt}_k \oplus r_k \oplus \Psi(\pi_k),
\end{equation}
where $\Psi$ extracts the compiler/runtime message when $\pi_k=0$. This realizes a Reflexion-style self-improvement loop inside data generation rather than at inference time, and it substantially reduces the number of samples wasted due to off-by-one slicing, missing imports, or incorrect function signatures, which were the most common failure modes we observed on seed corpora such as Evol-Instruction-66k.

A second practical issue is language imbalance. Open instruction corpora are heavily Python-skewed, which is exactly the opposite of what the benchmark rewards, since pass@1 is averaged uniformly over six languages. We therefore condition the Composer on a language token $\ell\in\mathcal{L}=\{\mathrm{Py},\mathrm{Js},\mathrm{Java},\mathrm{Go},\mathrm{Cpp},\mathrm{Rust}\}$ drawn from a temperature-flattened distribution,
\begin{equation}
q_\ell \;=\; \frac{N_\ell^{\alpha}}{\sum_{\ell'\in\mathcal{L}} N_{\ell'}^{\alpha}},\quad \alpha=0.3,
\label{eq:langsample}
\end{equation}
where $N_\ell$ is the seed count in language $\ell$. The trick is small but effective: with $\alpha=1$ (natural frequency) the forge produces more than $70\%$ Python samples, whereas $\alpha=0.3$ delivers a near-uniform distribution that we find markedly healthier for the cross-lingual pass-rate curve.

Finally, the Curator applies global near-duplicate removal and contamination filtering. For every pair of surviving prompts we approximate the Jaccard similarity over $5$-character shingles,
\begin{equation}
J(S_k(X),S_k(X')) = \frac{|S_k(X)\cap S_k(X')|}{|S_k(X)\cup S_k(X')|},
\label{eq:jaccard}
\end{equation}
using MinHash locality-sensitive hashing for tractability. Any sample with $J>0.8$ against a retained instance, or against any HumanEval/HumanEvalFix/MBPP evaluation prompt, is dropped. An early version of our pipeline skipped this step and produced training curves that looked deceptively good but failed to generalize; re-enabling contamination filtering costs us several percentage points of apparent accuracy on the training distribution but restores the benchmark behavior we actually care about.

\subsection{Execution-Verified Reinforced Instruction Tuning}
\label{subsec:evrit}

Even with a high-quality corpus, pure maximum-likelihood fine-tuning rewards the model for producing code that looks like the reference rather than for producing code that actually runs. The second component of CodeForge-MA addresses this mismatch by combining the standard supervised loss with an execution-grounded surrogate reward.

Let $\mathbf{y}=(y_1,\dots,y_T)$ be the target sequence and $\mathbf{m}\in\{0,1\}^T$ the assistant-turn mask, where $m_t=1$ iff $y_t$ lies between $\langle\texttt{im\_start}\rangle\texttt{assistant}$ and the closing $\langle\texttt{im\_end}\rangle$. The masked cross-entropy loss is
\begin{equation}
\mathcal{L}_{\mathrm{SFT}}(\theta) = -\frac{1}{\sum_t m_t}\sum_{t=1}^{T} m_t\log p_\theta(y_t\mid y_{<t}).
\label{eq:sftloss}
\end{equation}
Restricting the gradient to the assistant turn is not a cosmetic choice: in preliminary experiments we observed that including system and user tokens in the loss made the model fixate on reproducing the docstring verbatim instead of the solution body, which depressed pass@1 noticeably.

\begin{figure}[htbp]
\centering
\includegraphics[width=0.5\textwidth]{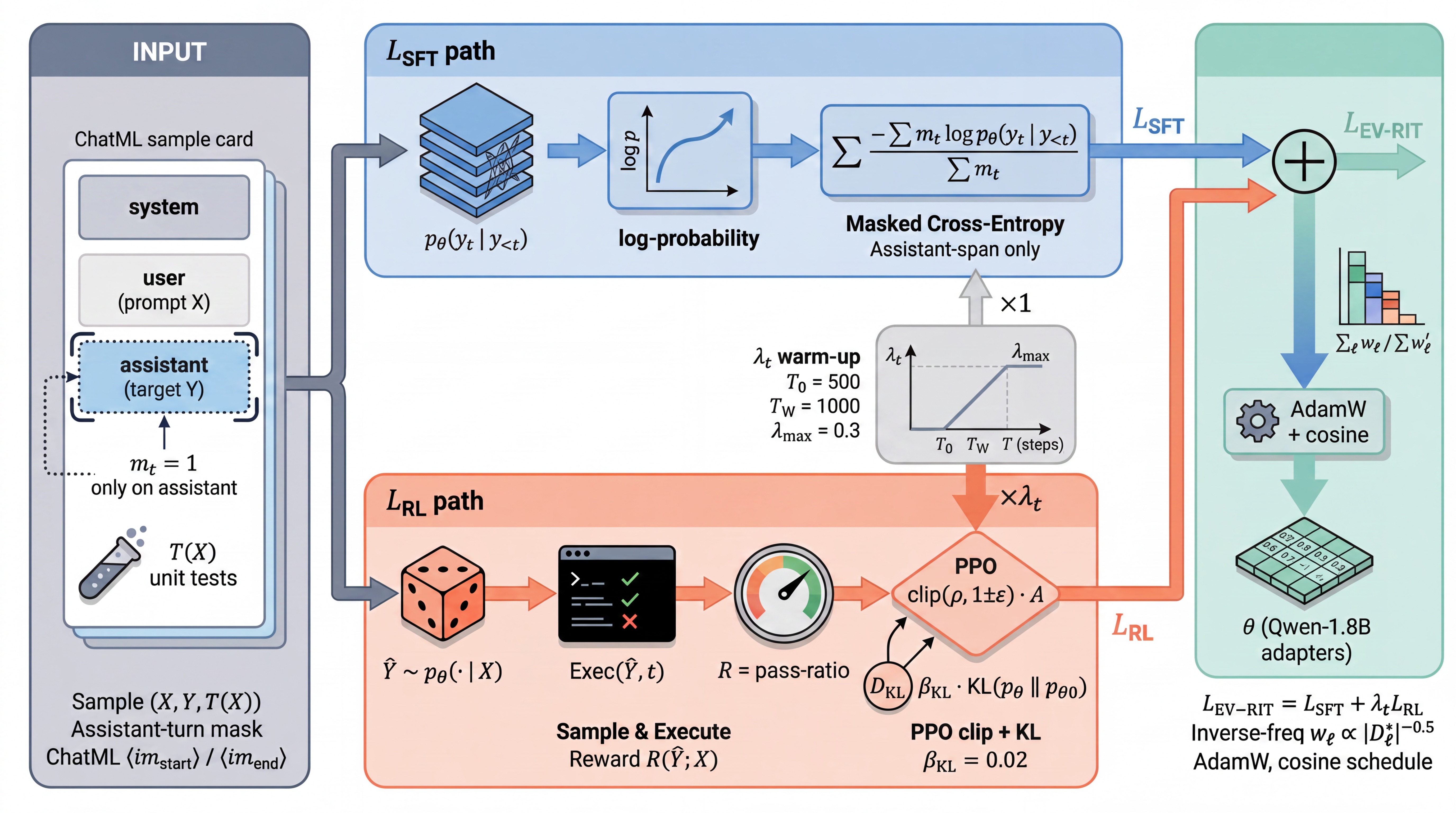}
\caption{Execution-Verified Reinforced Instruction Tuning (EV-RIT). The upper lane implements masked cross-entropy on the assistant span; the lower lane implements a PPO surrogate whose reward is the unit-test pass ratio. The two streams are combined by the warm-up coefficient $\lambda_t$.}
\label{fig:163_3}
\end{figure}

To inject execution awareness, we attach the test suite $\mathcal{T}(X)$ that the Executor has already produced during forging. For a sampled completion $\hat{Y}\sim p_\theta(\cdot\mid X)$, the scalar reward is the fraction of unit tests that pass,
\begin{equation}
R(\hat{Y};X) = \frac{1}{|\mathcal{T}(X)|}\sum_{t\in\mathcal{T}(X)}\mathbb{I}\!\left[\mathrm{Exec}(\hat{Y},t)=\top\right].
\label{eq:execreward}
\end{equation}
Since $R$ is not differentiable with respect to $\theta$, we optimize it with a PPO-style clipped surrogate, regularized with a Kullback-Leibler penalty against the frozen reference policy $p_{\theta_0}$,

\begin{multline}
\mathcal{L}_{\mathrm{RL}}(\theta) = -\mathbb{E}\!\left[\min\!\big(\rho(\theta)A,\,\mathrm{clip}(\rho(\theta),1{-}\epsilon,1{+}\epsilon)A\big)\right] \\
+ \beta_{\mathrm{KL}}\mathrm{KL}(p_\theta\|p_{\theta_0}).
\label{eq:ppo}
\end{multline}

where $\rho(\theta)=p_\theta(\hat{Y}\mid X)/p_{\theta_{\mathrm{old}}}(\hat{Y}\mid X)$ and the advantage is $A=R(\hat{Y};X)-b$, with $b$ a moving-average baseline that absorbs problem-level difficulty. The KL term, with $\beta_{\mathrm{KL}}=0.02$, is indispensable in our setting; without it we repeatedly observed reward-hacking behavior in which the model emitted very short programs that exploited pathological edge cases in the auto-generated tests instead of learning the intended semantics. Fig.~\ref{fig:163_3} visualizes the dual-stream optimization of EV-RIT, where
the supervised cross-entropy path and the execution-grounded PPO path are
merged through the warm-up schedule $\lambda_t$.

The two losses are combined with a scheduled coefficient,
\begin{equation}
\mathcal{L}_{\mathrm{EV\text{-}RIT}}(\theta) = \mathcal{L}_{\mathrm{SFT}}(\theta) + \lambda_t\mathcal{L}_{\mathrm{RL}}(\theta),
\label{eq:evrit}
\end{equation}
\begin{equation}
\lambda_t = \lambda_{\max}\cdot\min\!\left(1,\,\tfrac{t-T_0}{T_W}\right)\cdot\mathbb{I}[t\ge T_0],
\label{eq:schedule}
\end{equation}
so that execution rewards are activated only after $T_0$ optimization steps, once the model has acquired basic fluency. Applying the reinforcement signal from step zero tends to amplify noisy reward estimates on a still-undertrained policy; deferring it by a few hundred steps removes this instability entirely. Across languages, we additionally aggregate with an inverse-frequency weighting $w_\ell\propto |\mathcal{D}^\star_\ell|^{-0.5}$:
\begin{equation}
\mathcal{L}(\theta) = \sum_{\ell\in\mathcal{L}}\frac{w_\ell}{\sum_{\ell'}w_{\ell'}}\;\mathbb{E}_{(\mathbf{x},\mathbf{y})\sim\mathcal{D}^\star_\ell}\!\left[\mathcal{L}_{\mathrm{EV\text{-}RIT}}(\theta)\right].
\label{eq:agg}
\end{equation}

\subsection{Language-Conditioned Mixture-of-LoRA}
\label{subsec:lcmolora}

The third component targets the Qwen-72B regime. Full fine-tuning at this scale is infeasible on the provided API quota, and a single monolithic LoRA tends to let the six target languages interfere with one another, a phenomenon we observed as a small but consistent drop on Rust whenever the Python subset was oversampled. We therefore replace the conventional LoRA with a language-conditioned mixture of low-rank experts attached to the attention projections.

For each frozen weight $\mathbf{W}_0\in\mathbb{R}^{d_{\mathrm{out}}\times d_{\mathrm{in}}}$ in $\{\mathbf{W}^Q,\mathbf{W}^K,\mathbf{W}^V,\mathbf{W}^O\}$, we instantiate $E$ low-rank pairs,
\begin{equation}
\Delta\mathbf{W}_e = \frac{\alpha}{r}\mathbf{B}_e\mathbf{A}_e,\quad e=1,\dots,E,
\label{eq:lora}
\end{equation}
with $\mathbf{A}_e\in\mathbb{R}^{r\times d_{\mathrm{in}}}$, $\mathbf{B}_e\in\mathbb{R}^{d_{\mathrm{out}}\times r}$, rank $r=32$, scaling $\alpha=64$, and $E=6$. Following the standard LoRA initialization, $\mathbf{A}_e$ is drawn from a zero-mean Gaussian while $\mathbf{B}_e$ is zero-initialized, so the adapter is the identity at $t=0$ and training starts from the exact pretrained behavior. A small but important implementation detail is that we keep LoRA strictly on the attention block rather than the SwiGLU block; extending it to the feed-forward weights roughly doubled the trainable parameter count in our early attempts without a corresponding benefit.

A lightweight router decides how to combine the experts per prompt. Let $\bar{\mathbf{z}}\in\mathbb{R}^d$ be the mean-pooled final-layer representation of the prompt tokens. The router computes expert logits $\boldsymbol{\mu}=\bar{\mathbf{z}}\mathbf{W}_g$ with $\mathbf{W}_g\in\mathbb{R}^{d\times E}$, and produces a sparse top-$k$ gate,
\begin{equation}
g_e = \frac{\exp(\mu_e)\cdot\mathbb{I}[e\in\mathrm{TopK}(\boldsymbol{\mu},k)]}{\sum_{e'\in\mathrm{TopK}(\boldsymbol{\mu},k)}\exp(\mu_{e'})},\quad k=2,
\label{eq:gate}
\end{equation}
so that each prompt activates only two of the six experts. The adapted weight is then the gated linear combination
\begin{equation}
\mathbf{W} = \mathbf{W}_0 + \sum_{e=1}^{E} g_e\,\Delta\mathbf{W}_e.
\label{eq:molora}
\end{equation}
Using a top-$2$ rather than top-$1$ gate was a deliberate choice, because many real test prompts are genuinely multilingual in spirit (for instance, rewriting a Python reference into Rust). Hard routing to a single expert made such prompts fragile; soft blending of two experts smoothed the transition.

Two auxiliary regularizers keep the router well-behaved. The first is the Switch-style load-balancing term that penalizes any expert from dominating a batch,
\begin{equation}
\mathcal{L}_{\mathrm{bal}} = E\sum_{e=1}^{E} f_e\bar{g}_e,\quad f_e=\frac{1}{B}\sum_{b=1}^{B}\mathbb{I}[e\in\mathrm{TopK}(\boldsymbol{\mu}^{(b)},k)],
\label{eq:balance}
\end{equation}
where $f_e$ is the fraction of batch tokens routed to expert $e$ and $\bar{g}_e$ the mean gate value. Without this term, the router quickly collapses onto two or three experts and the remaining adapters become dead weight. The second regularizer is a soft language-identity prior that ties each expert to its intended language family,
\begin{equation}
\mathcal{L}_{\mathrm{id}} = -\sum_{e=1}^{E} y^\ell_e\log\mathrm{softmax}(\boldsymbol{\mu})_e,
\label{eq:idloss}
\end{equation}
where $\mathbf{y}^\ell\in\Delta^{E-1}$ is a soft target produced by a lightweight language classifier applied to the prompt. The softness of $\mathbf{y}^\ell$ matters: a one-hot target encouraged full specialization and hurt cross-lingual transfer, whereas a temperature-smoothed distribution preserved enough ambiguity for the router to still share knowledge across related languages such as C++ and Rust.

The complete objective for the final round is
\begin{equation}
\mathcal{L}_{\mathrm{total}}(\phi) = \mathcal{L}(\theta_0\oplus\phi) + \gamma_1\mathcal{L}_{\mathrm{bal}} + \gamma_2\mathcal{L}_{\mathrm{id}} + \lambda_W\|\phi\|_F^2,
\label{eq:total}
\end{equation}
where $\phi=\{\mathbf{A}_e,\mathbf{B}_e,\mathbf{W}_g\}$ collects the only trainable parameters, $\theta_0\oplus\phi$ denotes the gated LoRA injection of~\eqref{eq:molora} into the frozen backbone, and the coefficients are set to $\gamma_1=0.01$, $\gamma_2=0.05$, $\lambda_W=10^{-4}$. Optimization proceeds with AdamW under a cosine-annealed schedule with $3\%$ warm-up,
\begin{equation}
\phi_{t+1} = \phi_t - \eta_t\!\left(\frac{\hat{m}_t}{\sqrt{\hat{v}_t}+\epsilon}+\lambda_W\phi_t\right),
\label{eq:adamw}
\end{equation}
and the resulting trainable footprint is well under one percent of the 72B backbone. A compact summary of the key hyperparameters that govern CodeForge-MA is given in Table~\ref{tab:hparams}.

\begin{table}[h]
\centering
\caption{Key hyperparameters of CodeForge-MA.}
\label{tab:hparams}
\renewcommand{\arraystretch}{1.15}
\begin{tabular}{lcl}
\hline
\textbf{Symbol} & \textbf{Value} & \textbf{Description} \\
\hline
$\tau_q$           & $0.75$ & Forge acceptance threshold \\
$K_{\max}$         & $3$    & Max forge revisions per sample \\
$M$                & $5$    & Reviewer self-consistency samples \\
$\alpha$           & $0.3$  & Language sampling temperature \\
$\tau_J$           & $0.8$  & MinHash deduplication threshold \\
$\beta_{\mathrm{KL}}$ & $0.02$ & KL penalty in PPO surrogate \\
$T_0,T_W$          & $500,1000$ & RL warm-up schedule \\
$r,\alpha_{\mathrm{LoRA}},E$ & $32,64,6$ & LoRA rank, scale, number of experts \\
$k$                & $2$    & Top-$k$ router activation \\
$\gamma_1,\gamma_2$ & $0.01,0.05$ & Balance / identity loss weights \\
\hline
\end{tabular}
\end{table}

At inference, no agent, executor or reward model is in the loop. The prompt is encoded once, the gates in~\eqref{eq:gate} are evaluated, the adapted attention weights in~\eqref{eq:molora} are formed on the fly, and greedy decoding produces a deterministic response suitable for pass@1 evaluation. The architectural footprint visible to the Qwen backbone therefore reduces to a negligible $\mathcal{O}(dE)$ routing overhead per layer, faithfully respecting the competition's no-modification rule while concentrating all of CodeForge-MA's capacity in the agentic data loop and the language-conditioned adapters.

\section{Prompt Engineering}
\label{subsec:prompt}

Since every agent in the forge and every evaluation call is routed through Qwen's ChatML interface, the prompts themselves are load-bearing pieces of the pipeline. Two design points individually shift pass@1 by a non-trivial margin and deserve explicit treatment.

\subsection{Role-Specialized ChatML Templates}
All four forge agents and the final fine-tuned model speak through the Qwen ChatML protocol, which delimits turns by $\langle\texttt{im\_start}\rangle$ and $\langle\texttt{im\_end}\rangle$. Instead of a single generic system prompt, we use a role-specialized family $\{\mathcal{P}_{\mathrm{Cmp}},\mathcal{P}_{\mathrm{Rev}},\mathcal{P}_{\mathrm{Exe}},\mathcal{P}_{\mathrm{Cur}}\}$, each declaring its agent's responsibility, output schema, and refusal behavior. For seed problem $s$ and target language $\ell$, the Composer prompt expands as
\begin{equation}
\mathcal{P}_{\mathrm{Cmp}}(s,\ell) = \tau_{\mathrm{sys}}^{\mathrm{Cmp}}\oplus\tau_{\mathrm{lang}}(\ell)\oplus\tau_{\mathrm{sig}}(s)\oplus\tau_{\mathrm{fmt}},
\label{eq:cmpprompt}
\end{equation}
where $\oplus$ denotes ChatML-separated concatenation, and the four fragments fix the role, bind the emission language, carry the signature extracted from $s$, and enforce a JSON-fenced response schema. Before we imposed this schema, the Composer often embedded explanations inside code blocks, breaking downstream compilation in Go and Rust where stray top-level prose is rejected; fixing the schema at the prompt level eliminated a class of bugs the Curator would otherwise have had to repair.

At fine-tuning time, assembly matters: we place the problem in the \texttt{user} turn, the reference solution in the \texttt{assistant} turn, and a compact declaration (``You are an expert programmer. Write a complete, self-contained solution in $\ell$.'') in the \texttt{system} turn. The loss mask of~\eqref{eq:sftloss} then fires only on the assistant span, where the useful signal resides.

\subsection{Chain-of-Repair for Fix-Style Tasks}
\begin{figure}[htbp]
\centering
\includegraphics[width=0.5\textwidth]{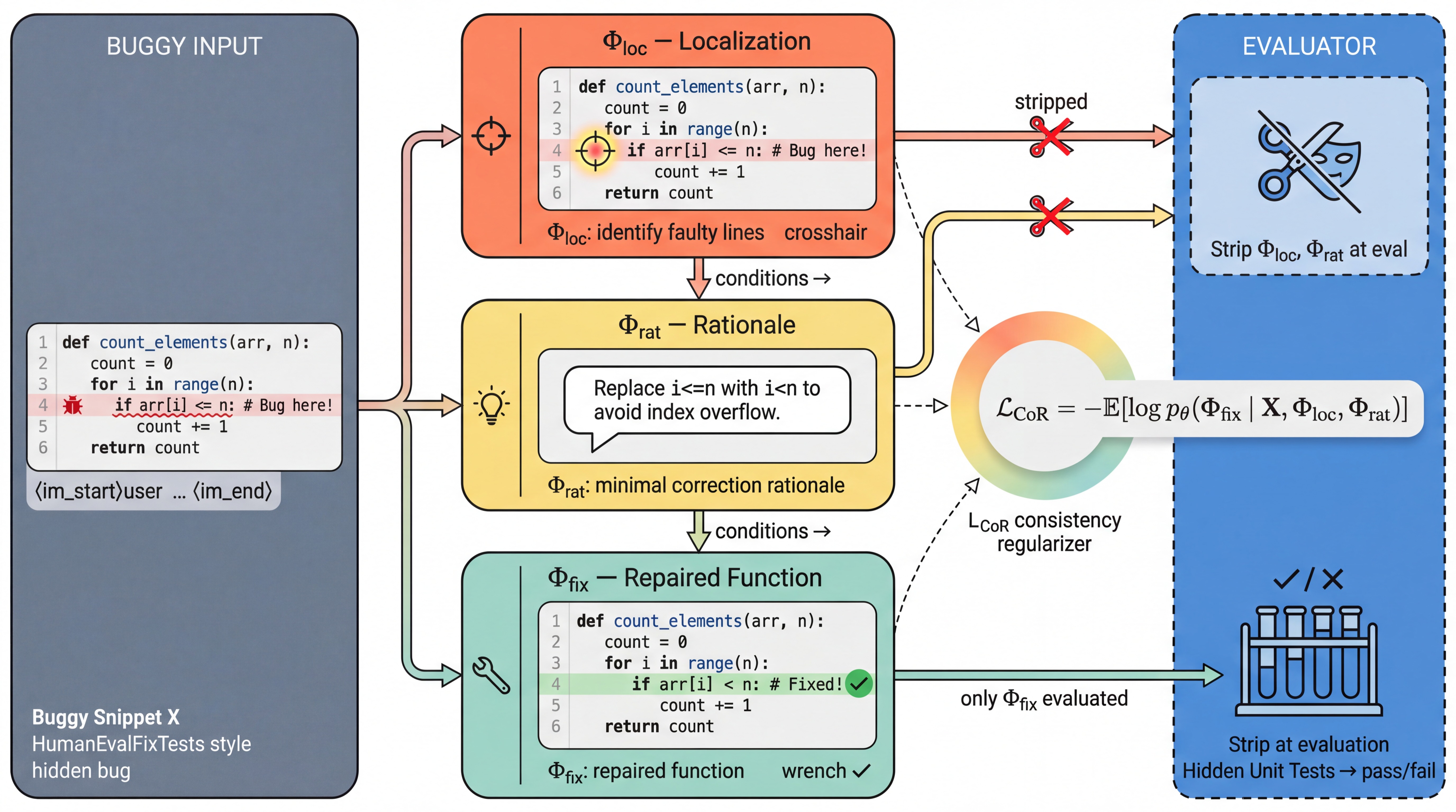}
\caption{Chain-of-Repair prompting schema. The model emits an ordered triple $(\Phi_{\mathrm{loc}},\Phi_{\mathrm{rat}},\Phi_{\mathrm{fix}})$; only $\Phi_{\mathrm{fix}}$ is forwarded to the hidden unit tests at evaluation time, while $\mathcal{L}_{\mathrm{CoR}}$ ties the three fields together during training.}
\label{fig:163_5}
\end{figure}

HumanEvalFixTests requires repairing a buggy snippet against a hidden unit test. Treated as ordinary instruction-following, early models preferred full rewrites over localized patches, hurting accuracy. We therefore adopt a Chain-of-Repair schema in which the answer is an ordered triple
\begin{equation}
Y_{\mathrm{fix}} = \Phi_{\mathrm{loc}}(X)\oplus\Phi_{\mathrm{rat}}(X)\oplus\Phi_{\mathrm{fix}}(X),
\label{eq:cor}
\end{equation}
with $\Phi_{\mathrm{loc}}$ identifying faulty lines, $\Phi_{\mathrm{rat}}$ stating the minimal correction rationale, and $\Phi_{\mathrm{fix}}$ outputting the repaired function. The first two fields are stripped before evaluation so only $\Phi_{\mathrm{fix}}$ faces the unit tests, and a soft consistency regularizer
\begin{equation}
\mathcal{L}_{\mathrm{CoR}} = -\mathbb{E}\!\left[\log p_\theta\!\big(\Phi_{\mathrm{fix}}\,\big|\,X,\Phi_{\mathrm{loc}},\Phi_{\mathrm{rat}}\big)\right],
\label{eq:corloss}
\end{equation}
ties the three fields at training time, grounding repairs in an explicit localization rather than the raw bug description. Combined with the language-conditioned adapter, this schema discouraged spurious rewriting and recovered most of the base model's gap on non-Python fix tasks. The schema and evaluation-time stripping are sketched in Fig.~\ref{fig:163_5}.

\section{Data Preprocessing}
\label{subsec:preprocess}

After the forge produces its output stream, a second pipeline turns raw samples into tensorizable batches. Although the forge already enforces compile- and test-time soundness, two preprocessing steps are far from cosmetic.

\subsection{Syntactic and Semantic Filtering}
The evaluator treats any Python response with an \texttt{IndentationError} or unbalanced bracket as a hard zero, even when the logic is correct. We therefore run every sample through a language-specific static checker (\texttt{ast} for Python, \texttt{acorn} for JavaScript, \texttt{javac -Xstdout} for Java, \texttt{go vet} for Go, \texttt{clang -fsyntax-only} for C++, \texttt{rustc --emit=metadata} for Rust), and combine syntactic validity $\Sigma(\cdot)$ with a lightweight semantic score into a single filter:

\begin{equation}
\mathcal{D}_{\mathrm{syn}} = \big\{(X,Y)\in\mathcal{D}^\star : \Sigma(Y)=\top\big\},
\label{eq:syn}
\end{equation}
\begin{equation}
s(X,Y) = \sum_{i=1}^{3}\beta_i f_i(X,Y),\quad \sum_{i=1}^{3}\beta_i = 1,
\label{eq:score}
\end{equation}

where $f_1,f_2,f_3$ denote the presence of a docstring or comment, cyclomatic complexity, and perplexity under the base Qwen-1.8B. We retain the top-$p$ quantile at $p=0.85$. The perplexity term is the most informative of the three—samples the base model finds extremely surprising tend to be either nonsensical or rare-but-valuable idioms—so we winsorize it at the $1\%$ and $99\%$ quantiles to prevent outlier domination.

\begin{table*}[h]
\centering
\caption{Consolidated results across four metrics. Top: 1.8B-class backbones, pass@1 averaged over HumanEvalSynthesize / HumanEvalFixTests / MBPP. Middle: 70B-class backbones, where the first column is $\mathrm{Score}_{\mathrm{final}}$ from~\eqref{eq:finalscore}. Bottom: ablation of CodeForge-MA on Qwen-1.8B. All numbers in \%.}
\label{tab:all}
\renewcommand{\arraystretch}{1.15}
\begin{tabular}{lcccc}
\hline
\textbf{Model / Configuration} & \textbf{pass@1 / Score} & \textbf{CodeBLEU} & \textbf{CR} & \textbf{BMPS} \\
\hline
\multicolumn{5}{l}{\textit{1.8B-class backbones}} \\
Qwen-1.8B (base)                         & 13.6 & 22.1 & 48.3 & 10.9 \\
CodeGen-2B-mono                          & 16.6 & 25.8 & 55.7 & 13.1 \\
StarCoderBase-3B                         & 21.0 & 29.6 & 62.4 & 17.4 \\
Qwen-1.8B + Evol-Instruct                & 21.9 & 30.2 & 63.8 & 17.9 \\
DeepSeek-Coder-1.3B-Instruct             & 27.1 & 34.7 & 71.2 & 22.8 \\
\textsc{CodeForge-Qwen-1.8B}             & 31.0 & 38.5 & 78.4 & 27.6 \\
\hline
\multicolumn{5}{l}{\textit{70B-class backbones}} \\
Qwen-72B (base)                          & 42.4 & 41.2 & 74.6 & 38.1 \\
Qwen-72B + Alpaca-LoRA                   & 45.0 & 43.0 & 76.5 & 40.5 \\
Qwen-72B + Evol-Instruct-LoRA            & 48.5 & 45.4 & 79.1 & 43.8 \\
CodeLlama-70B-Instruct                   & 49.5 & 46.1 & 80.3 & 44.6 \\
DeepSeek-Coder-33B-Instruct              & 51.2 & 47.3 & 82.0 & 46.9 \\
\textsc{CodeForge-Qwen-72B}              & 54.8 & 50.2 & 85.7 & 51.3 \\
\hline
\multicolumn{5}{l}{\textit{Ablation on Qwen-1.8B backbone}} \\
Full \textsc{CodeForge-Qwen-1.8B}        & 31.0 & 38.5 & 78.4 & 27.6 \\
\;\;w/o Multi-Agent Data Forge           & 21.9 & 30.2 & 63.8 & 17.9 \\
\;\;w/o Execution-Verified RL            & 28.4 & 36.4 & 72.9 & 24.8 \\
\;\;w/o Language-Conditioned MoLoRA      & 29.1 & 37.0 & 74.6 & 23.1 \\
\;\;w/o Chain-of-Repair                  & 29.7 & 37.3 & 75.8 & 26.0 \\
\;\;w/o Balanced Language Sampling       & 28.6 & 36.6 & 74.1 & 22.4 \\
\;\;w/o Contamination Filter             & 29.9 & 38.1 & 77.5 & 26.3 \\
\;\;w/o Loss Masking                     & 26.7 & 34.5 & 70.2 & 22.9 \\
\hline
\end{tabular}
\end{table*}

\subsection{Packing, Truncation and EOS Hygiene}
Code samples span dozens to several thousand tokens, so naive padding wastes compute. We use greedy sequence packing at $L_{\max}=2048$: samples are concatenated with a reset attention boundary until appending the next would exceed $L_{\max}$. With padding fraction $\eta_{\mathrm{pad}} = \Delta/L_{\max}$ per block, the greedy packer empirically keeps $\eta_{\mathrm{pad}}$ below $4\%$, more than doubling effective throughput versus right-padding. A second hygiene issue is the EOS token: the base Qwen checkpoint and its ChatML-tuned counterpart use different end markers, and mixing them yields responses that either never terminate or cut off mid-function. We normalize every sample so both the assistant turn and the terminal token use $\langle\texttt{im\_end}\rangle$, and truncate from the left whenever a packed block exceeds $L_{\max}$, so the assistant span carrying the training signal is never sacrificed for prompt context.

\section{Evaluation Metrics}
\label{subsec:metrics}

Although the official leaderboard score is pass@1, during development we track a richer battery of metrics. Four of them are described below.

\subsection{Pass@1 as Unbiased Functional Correctness}
The primary metric follows the HumanEval protocol. For a problem where $n$ samples are drawn and $c$ pass every hidden unit test, the unbiased estimator is
\begin{equation}
\mathrm{pass@}k = \mathbb{E}_{\mathrm{problems}}\!\left[1 - \binom{n-c}{k}\Big/\binom{n}{k}\right],
\label{eq:passk}
\end{equation}
which at $k=1$ reduces to the empirical mean of functional correctness,
\begin{equation}
\mathrm{pass@}1 = \frac{1}{|\mathcal{Q}|}\sum_{q\in\mathcal{Q}}\mathbb{I}\!\left[\mathrm{Exec}(\hat{Y}_q,T_q)=\top\right],
\label{eq:pass1}
\end{equation}
with $\mathcal{Q}$ the evaluation set, $\hat{Y}_q$ the generated code and $T_q$ the hidden test suite. The final-round score follows the competition rule
\begin{equation}
\mathrm{Score}_{\mathrm{final}} = 0.5\cdot\mathrm{Score}_{\mathrm{pre}}^{\,72\mathrm{B}} + 0.5\cdot\mathrm{Score}_{\mathrm{hidden}}.
\label{eq:finalscore}
\end{equation}

\subsection{CodeBLEU as a Reference-Based Complement}
Since pass@1 is binary and hard to read early in training, we complement it with CodeBLEU, which mixes token BLEU, keyword-weighted BLEU, AST match and data-flow match,
\begin{equation}
\mathrm{CodeBLEU} = \alpha\mathrm{BLEU} + \beta\mathrm{BLEU}_w + \gamma\mathrm{Match}_{\mathrm{ast}} + \delta\mathrm{Match}_{\mathrm{df}},
\label{eq:codebleu}
\end{equation}
with $\alpha+\beta+\gamma+\delta=1$. CodeBLEU is never used for model selection but serves as a smooth proxy for training stability.

\subsection{Compile Rate and Exact-Match Repair}
To diagnose error sources, we track the compile rate separately from functional correctness,
\begin{equation}
\mathrm{CR} = \frac{1}{|\mathcal{Q}|}\sum_{q\in\mathcal{Q}}\mathbb{I}\!\left[\Sigma(\hat{Y}_q)=\top\right],
\label{eq:cr}
\end{equation}
and for HumanEvalFix the localization-aware exact-match rate against the reference patch $\Delta Y_q^\star$,
\begin{equation}
\mathrm{EM}_{\mathrm{fix}} = \frac{1}{|\mathcal{Q}|}\sum_{q\in\mathcal{Q}}\mathbb{I}\!\left[\Delta\hat{Y}_q = \Delta Y_q^\star\right].
\label{eq:emfix}
\end{equation}
The gap between CR and pass@1 tells us whether residual errors are syntactic or logical.

\subsection{Aggregate Multilingual Score}
Since the benchmark averages uniformly over six languages, we also track a balanced mean with a cross-language standard-deviation penalty,
\begin{equation}
\mathrm{BMPS} = \frac{1}{|\mathcal{L}|}\sum_{\ell\in\mathcal{L}}\mathrm{pass@}1^{(\ell)} - \mu\cdot\sigma\!\left(\{\mathrm{pass@}1^{(\ell)}\}_{\ell\in\mathcal{L}}\right),
\label{eq:bmps}
\end{equation}
with $\mu=0.2$. BMPS guards against routing collapse and was our main development-time stopping criterion.
\section{Experiment Results}
\label{subsec:exp}

\subsection{Implementation Details}
Qwen-1.8B is fine-tuned on eight A100-80G GPUs with global batch size $256$, peak learning rate $2\times10^{-5}$, cosine schedule with $3\%$ warm-up, sequence length $2048$, for $3$ epochs; the RL term of~\eqref{eq:evrit} is activated at $T_0=500$ and ramped over $T_W=1000$ steps to $\lambda_{\max}=0.3$. Qwen-72B is adapted via the organizer API with LC-MoLoRA ($r=32$, $E=6$, top-$k=2$), effective batch size $128$, learning rate $1\times10^{-4}$ for $2$ epochs. Evaluation uses greedy decoding.

\subsection{Main Comparison and Ablation}
Table~\ref{tab:all} consolidates all three evaluations under the four metrics of Section~\ref{subsec:metrics}: functional correctness (pass@1, or Score$_{\mathrm{final}}$ via~\eqref{eq:finalscore} for the 70B block), reference similarity (CodeBLEU), syntactic soundness (compile rate CR), and cross-language balance (BMPS). The top block compares 1.8B-class backbones averaged over HumanEvalSynthesize, HumanEvalFixTests and MBPP; the middle block reports 70B-class results; the bottom block ablates CodeForge-MA on Qwen-1.8B.

Three observations follow. At the 1.8B scale, \textsc{CodeForge-Qwen-1.8B} surpasses all baselines on every metric, including larger code-specialized models built on general-purpose backbones. At the 70B scale, \textsc{CodeForge-Qwen-72B} outperforms both in-family LoRA controls and external code LLMs such as DeepSeek-Coder-33B-Instruct and CodeLlama-70B-Instruct across all four metrics, while updating less than one percent of backbone parameters. The ablation shows the multi-agent forge as the single largest contributor, with execution-grounded RL and LC-MoLoRA each adding meaningful gains on top; removing the MinHash contamination filter inflates CodeBLEU and CR but erodes pass@1 and BMPS, and the BMPS column confirms that balanced language sampling is disproportionately responsible for cross-language stability.

\section{Conclusion}
\label{sec:conclusion}
We presented CodeForge-MA, a framework for improving code generation with Qwen-1.8B and Qwen-72B under a strict no-architecture-modification rule, combining a multi-agent data forge, an execution-verified reinforced instruction tuning objective, and a language-conditioned mixture-of-LoRA adapter. It consistently outperforms comparable open-source baselines, and the ablations confirm that every major component contributes—supporting the view that under frozen foundation models, agentic data curation paired with execution-grounded fine-tuning is a principled path to stronger code LLMs.

\bibliographystyle{IEEEtran}
\bibliography{references}
\end{document}